\documentclass[11pt]{article}
\usepackage[final]{acl}
\usepackage{times}
\usepackage{latexsym}
\usepackage[T1]{fontenc}
\usepackage[utf8]{inputenc}
\usepackage{microtype}
\usepackage{inconsolata}
\usepackage{microtype}
 \usepackage{xcolor}
 \usepackage{adjustbox}
 \usepackage{array}
\usepackage{tabularx}
\usepackage{amsmath}
\usepackage{amsfonts}
\usepackage{graphicx}
\usepackage{float}
\usepackage{placeins}
\usepackage{subcaption}
\usepackage{xcolor}
\usepackage{booktabs}
\usepackage{multirow}
\usepackage[table]{xcolor}
\usepackage{enumitem}
\usepackage[final]{changes}
\usepackage{xspace}


\title{Cascade Failure Attribution for Multi-Agent Systems}

\author{
Jiayi Zhang\textsuperscript{1},
Zexin Wang\textsuperscript{1},
Degang Sun\textsuperscript{1},
Changhua Pei\textsuperscript{1},
Fei Sun\textsuperscript{2},
Gaogang Xie\textsuperscript{1},
Jingjing Li\textsuperscript{1}
\\
\textsuperscript{1}Computer Network Information Center, Chinese Academy of Sciences
\\
\textsuperscript{2}Institute of Computing Technology, Chinese Academy of Sciences
\\
\texttt{zhangjiayi26@mails.ucas.ac.cn}
\\
\texttt{\{wangzexin,dgsun,chpei,xie,ljj\}@cnic.cn},
\texttt{sunfei@ict.ac.cn}
}

\begin{document}

\maketitle

\begin{abstract}

Large language model (LLM)-based agents have shown strong potential in solving complex tasks through multi-step reasoning, yet they remain vulnerable to execution failures. Accurate failure attribution is therefore critical for improving agent reliability. Existing topology- and spectrum-based methods exploit trajectory structures but often overlook fine-grained semantics, while LLM-based attribution methods capture semantic cues but suffer from long-context degradation over lengthy trajectories. To address these challenges, we propose \textsc{DuoTrace}, a plug-and-play detection filter for LLM-based failure attribution. \textsc{DuoTrace} follows a detect-before-attribute paradigm: it first detects anomalous executions and then supplies focused trajectory evidence to downstream LLM-based attribution methods. For effective VAE-based anomaly detection on agent trajectories,
\textsc{DuoTrace} integrates dual-view semantic-structural node
representations, a Tree-LSTM-based trajectory encoder, and data augmentation through outcome-verified LLM replay and root-to-node path extraction to handle heterogeneous nodes, hierarchical execution structures, and
limited normal execution data. Experiments with six LLM-based attribution baselines show that \textsc{DuoTrace} improves agent-level and step-level attribution accuracy by 11.2\% and 7.7\%, respectively.
\end{abstract}

\section{Introduction}

Large language models (LLMs) have empowered autonomous agents to tackle complex scenarios through reasoning, planning, and environmental interaction \cite{yao2023tree,yao2023react}. As modern agent systems increasingly shift towards long-horizon tasks, their execution typically unfolds as a complex, hierarchical tree of highly coupled operations. However, this multi-step architecture introduces a critical vulnerability:  a subtle error at an early stage goes unnoticed, cascading through the execution branches and corrupting downstream decisions. This delayed manifestation of errors not only derails the final objective but also incurs substantial computational overhead and wasted inference time. Failure attribution (FA) identifies the agent and execution step responsible for a failed outcome, providing actionable evidence for debugging multi-agent systems \cite{zhang2025who,ge2025famas}.

\begin{figure}[t]
    \centering
    \begin{subfigure}[t]{0.48\columnwidth}
        \centering
        \includegraphics[width=\linewidth]{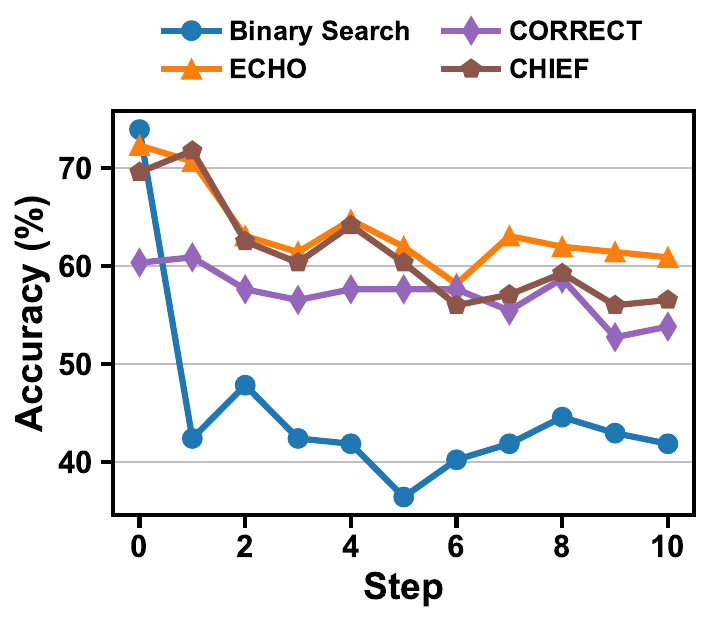}
        \caption{Agent-level accuracy.}
        \label{fig:agent-level}
    \end{subfigure}
    \hfill
    \begin{subfigure}[t]{0.48\columnwidth}
        \centering
        \includegraphics[width=\linewidth]{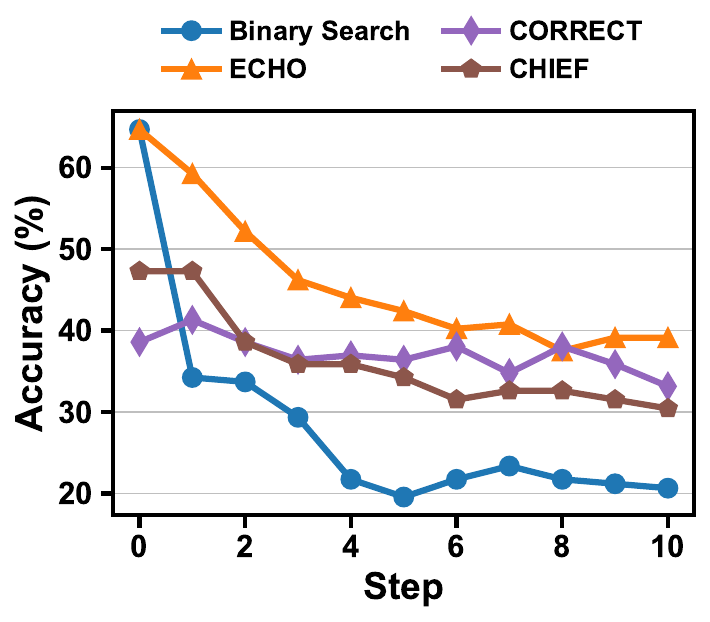}
        \caption{Step-level accuracy.}
        \label{fig:step-level}
    \end{subfigure}
    \caption{Attribution accuracy as a function of input trajectory length.}
    \label{fig:intro}
\end{figure}

Existing failure attribution methods exploit different
sources of evidence, including execution spectra
\citep{ge2025famas} and causal dependencies
\citep{ma2025cdcmas}.
However, execution statistics alone may not adequately
capture semantic failures such as hallucinations.
LLM-as-a-Judge methods can incorporate textual semantics
alongside structural context when assessing execution
trajectories. However, Figure~\ref{fig:intro} shows that attribution
accuracy tends to decrease as input trajectories become longer.
Providing more context does not necessarily improve
LLM-based judgments: longer trajectories can introduce
redundant or irrelevant information that obscures
failure-relevant evidence, making it harder to distinguish
the initiating error from its downstream consequences.

To reduce the distraction from redundant or irrelevant
execution details, we propose a detect-before-attribute
paradigm: identify suspicious execution nodes before
invoking an LLM judge, and retain these candidates
with their local context as focused attribution evidence.
We adopt a variational autoencoder (VAE) to learn normal
execution patterns and use node-level reconstruction
errors to prioritize candidates without node-level
failure labels during detector training.
Applying this approach to agent trajectories presents
three challenges:

\noindent(\textit{i}) \textbf{Heterogeneous anomaly manifestations.}
Execution anomalies can manifest as semantic inconsistencies,
disrupted invocation dependencies, or mismatches between
node content and its structural context.
This heterogeneity makes anomalies difficult to characterize
through either textual or structural signals alone.

\noindent(\textit{ii}) \textbf{Hierarchical execution dependencies.}
Task decomposition and nested tool calls create dependencies
across execution branches. A node's behavior must therefore
be assessed in relation to its invocation context, which
may not be captured by its position in a linear sequence.

\noindent(\textit{iii}) \textbf{Limited normal data.}
High-quality normal trajectories are scarce and may cover
only a narrow range of valid execution paths, making it
difficult for a VAE to distinguish legitimate behavioral
variation from anomalous execution.

To address these challenges, we propose \textsc{DuoTrace},
a dual-view VAE filter for LLM-based failure attribution. The main contributions of this work are summarized as follows:
\begin{itemize}[
    leftmargin=*,
    itemsep=2pt,
    parsep=0pt,
    topsep=3pt,
    partopsep=0pt
]
    \item We propose \textsc{DuoTrace}, a plug-and-play
    detect-before-attribute framework that selects suspicious
    execution nodes and their local context to provide focused
    evidence for downstream LLM-based failure attribution.

    \item To address heterogeneous anomalies and hierarchical
    execution dependencies, we combine contextual semantic
    representations with invocation-based Tree-LSTM aggregation.
    A VAE models normal execution patterns, while trajectory-normalized reconstruction errors and parent-relative changes
    prioritize suspicious nodes.

\item To address normal-data scarcity, we combine
outcome-verified LLM replay with root-to-node path
extraction to cover diverse successful paths and
intermediate states.

    \item Experiments with six downstream attribution methods
    on Who\&When show average absolute improvements of
    11.2\% and 7.7\% in overall agent and step accuracy,
    respectively, alongside reduced token consumption
    and inference time in the evaluated efficiency settings.
\end{itemize}

\section{Related Work}

\subsection{LLM-based Multi-Agent Systems}
The rapid evolution of LLMs has enabled autonomous
multi-agent systems (MAS) to handle complex tasks
\citep{wang2024survey, chen2024survey}. MAS frameworks support role-playing collaboration, as exemplified by CAMEL~\citep{li2023camel}, and SOP-driven workflows, as exemplified by MetaGPT~\citep{hong2024metagpt}. Recent advances further optimize routing
\citep{yue2025masrouter}, prompts \citep{khattab2023dspy, yuksekgonul2024textgrad}, and topologies \citep{zhuge2024gptswarm, zhang2024gdesigner}. However, as MAS scale and integrate diverse tools \citep{qin2023toollm, schick2023toolformer}, they exhibit significant structural fragility \citep{cemri2025why}. The intricate inter-dependencies dictate that minor upstream hallucinations can seamlessly propagate, culminating in catastrophic task failures \citep{zhu2025where}. Thus, effectively debugging these vulnerable systems has become a critical frontier \citep{deshpande2025trail}.

\subsection{Failure Attribution in Multi-Agent Systems}

To mitigate cascading errors, FA aims to identify the exact error step and the responsible agent. While early benchmarks primarily evaluated overall task success \citep{huang2023mlagentbench, yoran2024assistantbench} or collaboration dynamics \citep{zhu2025multiagentbench, ruan2025swarmbench}, recent studies have formalized FA as a distinct challenge. For example, the \textit{Who\&When} benchmark \citep{zhang2025who} annotates failure trajectories to assess diagnostic performance. Such benchmarks are further supported by data synthesis pipelines that use fault injection, such as AgenTracer \citep{zhang2025agentracer} and Aegis \citep{kong2025aegis}. Existing FA methods include spectrum-based analysis,
causal inference, and LLM-based evaluation.

\textbf{Spectrum-Based and Causal FA.}
FAMAS~\citep{ge2025famas} combines trajectory replay,
abstraction, and spectrum analysis to rank suspicious
agent actions. \citet{ma2025cdcmas} propose a causal
attribution framework that combines performance
dependency modeling with Shapley values for agent-level
attribution and uses CDC-MAS for step-level causal
discovery, incorporating features such as semantic
consistency. \textsc{DuoTrace} learns normal execution
patterns to prioritize suspicious nodes and provide
focused evidence for downstream attribution.

\textbf{LLM-as-a-Judge for FA.} To capture these semantic nuances, recent studies apply the LLM-as-a-Judge paradigm \citep{li2024llms, kamoi2024realmistake} to MAS debugging, similar to process-supervised evaluation \citep{zheng2024processbench, lightman2023letsverify, li2024process}. These approaches are either prompt-based diagnostic frameworks
using hierarchical contexts and error caches (e.g., ECHO
(Banerjee et al., 2025), CORRECT (Yu et al., 2025)) or
end-to-end fine-tuned models that directly predict failure steps
(e.g., AgenTracer (Zhang et al., 2025a)). Despite their semantic capabilities, LLM-based evaluators
incur substantial inference costs when processing lengthy
MAS logs.
Long-context studies further show that LLMs can struggle
to use relevant information located in the middle of
the input~\citep{liu2024lost}, motivating the selection
of focused evidence for attribution.

% === 图 1：总架构图 (跨双栏，放在页面顶部) ===
\begin{figure*}[t]
    \centering
    \includegraphics[width=\textwidth]{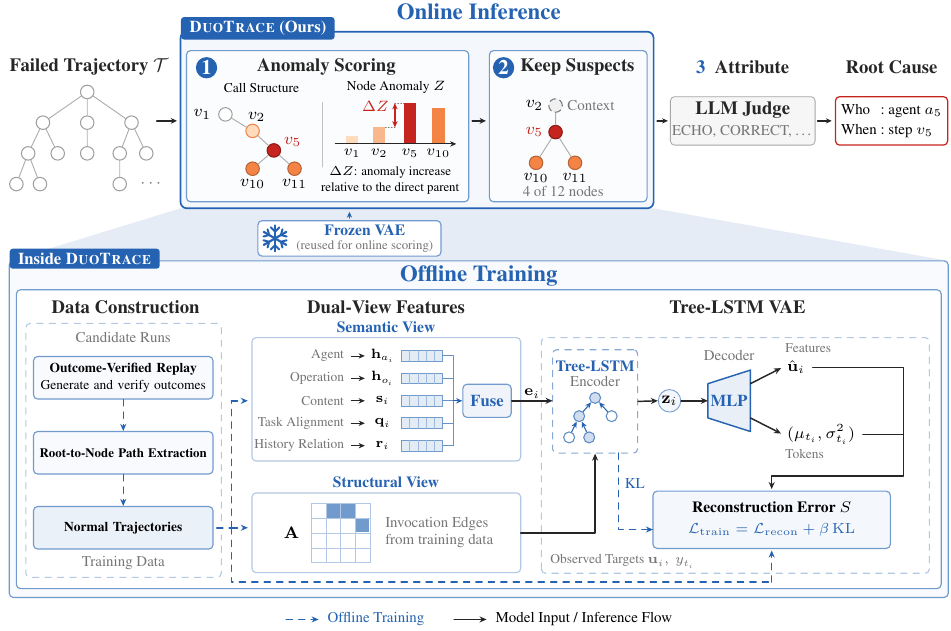} 
    \caption{The overall architecture of \textsc{DuoTrace}.}
    \label{fig:architecture}
\end{figure*}

\section{Motivation and Problem Formulation}
\label{sec:problem_formulation}

Multi-agent trajectories combine complex invocation
dependencies with verbose text. Feeding full trajectories
to LLM evaluators can obscure failure-relevant evidence
\citep{liu2024lost, li2025prompt}, while structural analysis
alone may overlook semantic failures.
We therefore introduce anomaly-based candidate selection
before LLM-based failure attribution.
\textsc{DuoTrace} learns normal execution patterns and
uses deviations to prioritize suspicious steps.
These anomaly scores guide candidate selection rather
than establish causal responsibility; downstream LLM
judges examine the retained candidates and their local
context for final attribution.

\subsection{Problem Formulation}

\textbf{Multi-Agent Execution Trajectory.}
Multi-agent execution traces contain both structural dependencies
and textual semantics \citep{wu2023autogen, hong2024metagpt}.
We represent a trajectory as a rooted directed tree
$\mathcal{T} = (\mathcal{V}, \mathcal{E})$,
using structural unrolling for non-tree traces.
Each node $v_i \in \mathcal{V}$ represents an execution step
$v_i = \langle a_i, o_i, x_i \rangle$,
where $a_i \in \mathcal{A}$ identifies the acting agent,
$o_i \in \mathcal{O}$ denotes the operator or tool,
and $x_i$ contains the textual payload.
Here, $\mathcal{A}$ and $\mathcal{O}$ are the agent and operator
sets, respectively.
Edges in $\mathcal{E}$ encode parent--child invocation
dependencies in this representation.

\textbf{Failure Attribution Task.}
Given a failed trajectory $\mathcal{T}$ and its task query,
single-root-cause attribution aims to identify the responsible
step $v^* \in \mathcal{V}$ and its acting agent
\citep{zhang2025who}.
\textsc{DuoTrace} supports this task through candidate filtering:
it assigns each node a score $R(v_i)$ and retains
$K = \lceil 0.2N \rceil$ nodes (i.e., the top $20\%$ of the
$N$ nodes in the trajectory), together with their parents and
direct children, to form an evidence set
$\mathcal{V}_{sub} \subseteq \mathcal{V}$.
The evidence and task query are then provided to downstream
LLM judges for final attribution
\citep{yu2025correct, banerjee2025echo}.
For multi-cause settings, we evaluate candidate-ranking quality
rather than assume a unique responsible step.

\section{Methodology}
\label{sec:methodology}

As shown in Figure~\ref{fig:architecture}, \textsc{DuoTrace}
operates in two stages.
During offline training, we construct augmented normal
trajectories through outcome-verified replay and root-to-node
path extraction, and train a Tree-LSTM VAE on semantic
and structural features to learn normal execution patterns.
During online attribution, the trained model scores nodes
in a failed trajectory using reconstruction errors and
parent-relative score increments.
We then select suspicious nodes and retain their local
context as focused evidence for downstream LLM-based
failure attribution.

\subsection{Reliable Normal Data Construction}
\label{subsec:data_augmentation}

\textbf{Outcome-Verified Replay.}
We collect source tasks from multiple benchmarks,
including SWE-bench~\citep{jimenez2024swebench},
VitaBench~\citep{he2025vitabench},
and WebArena~\citep{zhou2024webarena}.
Using only the original query and available context,
an LLM generates diverse candidate execution trajectories.
We verify their outcomes against reference answers or
environment success criteria and retain only those
that pass verification.
Reference answers are used solely for this
post-generation check and are never provided to the
trajectory generator.
Finally, we consolidate near-duplicate trajectories
based on execution structure and content similarity,
while preserving distinct successful paths.

\textbf{Root-to-Node Path Extraction.}
We extract paths from the root to each node in verified
successful trajectories and combine these paths with
complete trajectories for training.
The extracted paths preserve ancestor--descendant
invocation dependencies and provide training examples
at different execution depths.

\subsection{Dual-View Features}
\label{subsec:feature_representation}

For each node $v_i \in \mathcal{V}$ of the execution tree
$\mathcal{T}$, we construct a semantic representation
and aggregate node information along invocation edges
to capture structural dependencies.

\textbf{Semantic View.}
A Sentence Transformer $\phi$~\citep{reimers2019sentence}
maps the agent identity $a_i$ and invoked operation $o_i$
to embeddings $\mathbf{h}_{a_i}=\phi(a_i)$ and
$\mathbf{h}_{o_i}=\phi(o_i)$, capturing the functional
role of each execution step.
To characterize its behavior in context, we further
encode the current textual content, its alignment with
the task query, and its semantic relations to ancestor
steps:
\begin{equation}
\begin{aligned}
    \mathbf{s}_i &= \phi(x_i),\\
    \mathbf{q}_i &= \rho(Q,x_i),\\
    \mathbf{r}_i &=
    \frac{1}{|\mathcal{A}_i|}
    \sum_{v_j\in\mathcal{A}_i}\rho(x_j,x_i),
    \qquad |\mathcal{A}_i|>0,
\end{aligned}
\end{equation}
where $Q$ is the initial query, $\mathcal{A}_i$ contains the ancestors of $v_i$, and $\rho$ is a pretrained text-pair encoder whose hidden representations describe semantic
relations. For the root, we set $\mathbf{r}_i=\mathbf{0}$.

These features capture current content, task alignment,
and history relations beyond agent and operation identities.

\textbf{Semantic Fusion.}
Learned affine projections $P_s$, $P_q$, and $P_r$ align the contextual features before concatenation. We fuse them and construct the final node input:
\begin{equation}
\label{eq:main_definition}
\begin{aligned}
    \mathbf{h}_i^{\mathrm{sem}}
    &=
    \mathbf{W}_f
    [P_s(\mathbf{s}_i)\oplus
     P_q(\mathbf{q}_i)\oplus
     P_r(\mathbf{r}_i)]
    +\mathbf{b}_f,\\
    \mathbf{e}_i
    &=
    \mathbf{W}_{\mathrm{in}}
    [\mathbf{h}_{a_i}\oplus
     \mathbf{h}_{o_i}\oplus
     \mathbf{h}_i^{\mathrm{sem}}]
    +\mathbf{b}_{\mathrm{in}},
\end{aligned}
\end{equation}
where $\oplus$ denotes concatenation. This integrates contextual relations with the node's
functional role before structural encoding.

\textbf{Structural View.}
The adjacency matrix $\mathbf{A}$ records invocation dependencies, with $A_{ij}=1$ when $v_i$ directly invokes $v_j$. Node features provide semantic information, while these edges determine which states are aggregated by the Tree-LSTM.

\subsection{Tree-LSTM VAE}
\label{subsec:vae_architecture}

\textbf{Tree-Structured Encoding.}
A Child-Sum Tree-LSTM~\citep{tai2015improved} combines each node feature with its children's
hidden and cell states:
\begin{equation}
(\mathbf{g}_i,\mathbf{c}_i)
=
\operatorname{Tree-LSTM}\!\left(
\mathbf{e}_i,
\{(\mathbf{g}_j,\mathbf{c}_j)\}_{v_j\in C(v_i)}
\right),
\end{equation}
where $C(v_i)$ denotes the direct children. The resulting state integrates contextual semantics
with the observed subtree. Learned projections of $\mathbf{g}_i$ parameterize
the mean and diagonal covariance of
$q_\theta(\mathbf{z}_i\mid\mathbf{g}_i)$. We sample the latent variable through reparameterization and regularize it with a standard Gaussian prior.

\textbf{Multi-Objective Reconstruction.}
An MLP decoder reconstructs the role, operation, content,
task-relation, and history-relation features.
Inspired by the joint modeling of invocation structure
and time consumption in TraceVAE~\citep{xie2023tracevae},
we additionally predict a Gaussian distribution over
per-node token consumption in agent trajectories.
The two errors are
\begin{equation}
\begin{aligned}
    E_i^{\mathrm{rec}}
    &=
    \sum_{\mathbf{u}\in\mathcal{B}_i}
    \lambda_{\mathbf{u}}
    \operatorname{MSE}(\mathbf{u},\hat{\mathbf{u}}),\\
    E_i^{\mathrm{tok}}
    &=
    -\log\mathcal{N}
    (y_{t_i};\mu_{t_i},\sigma_{t_i}^{2}),
\end{aligned}
\end{equation}
where $\mathcal{B}_i$ contains the reconstruction targets
$\mathbf{h}_{a_i}$, $\mathbf{h}_{o_i}$, $\mathbf{s}_i$,
$\mathbf{q}_i$, and $\mathbf{r}_i$, excluding $\mathbf{r}_i$
for the root.
$\hat{\mathbf{u}}$ and $\lambda_{\mathbf{u}}$ denote the
reconstruction of $\mathbf{u}$ and its loss weight.
$y_{t_i}$ is the observed node-level token count, with
decoder-predicted mean $\mu_{t_i}$ and variance $\sigma_{t_i}^{2}$.

We minimize the following objective, averaged over
normal training nodes:
\begin{equation}
\begin{aligned}
    \mathcal{L}_i
    =&\;
    \mathbb{E}_{q_\theta}
    [E_i^{\mathrm{rec}}+\lambda_tE_i^{\mathrm{tok}}]\\
    &+\beta D_{\mathrm{KL}}
    \left(
    q_\theta(\mathbf{z}_i\mid\mathbf{g}_i)
    \,\|\,\mathcal{N}(\mathbf{0},\mathbf{I})
    \right).
\end{aligned}
\end{equation}

In Figure~\ref{fig:architecture},
$\mathcal{L}_{\mathrm{recon}}$ denotes the posterior
expectation of the combined feature reconstruction
and weighted token NLL losses, averaged over normal
training nodes. Adding the corresponding mean KL
regularizer gives $\mathcal{L}_{\mathrm{train}}$.

\subsection{Anomaly Scoring}
\label{subsec:scoring_mechanism}

\textbf{Trajectory-Level Score Normalization.}
At inference time, we decode the posterior mean
and combine the feature reconstruction error with
the weighted token NLL to obtain $S(v_i)$,
shown as Reconstruction error $S$ in
Figure~\ref{fig:architecture}.
The KL term is used only during training.
We then standardize $S(v_i)$ within each trajectory:
\begin{equation}
\begin{aligned}
    S(v_i)
    &=E_i^{\mathrm{rec}}+\lambda_tE_i^{\mathrm{tok}},\\
    Z(v_i)
&=\frac{S(v_i)-m_{\mathcal{T}}}
{s_{\mathcal{T}}+\varepsilon}.
\end{aligned}
\end{equation}
Here, $m_{\mathcal{T}}$ and $s_{\mathcal{T}}$ are
the mean and population standard deviation of all
node scores within trajectory $\mathcal{T}$,
and $\varepsilon>0$ ensures numerical stability.
Thus, $Z(v_i)$ measures within-trajectory deviation.

\textbf{Parent-Relative Mutation Scoring.}
We measure the change in the normalized anomaly score
along each invocation edge:
\begin{equation}
    \Delta Z(v_i)=Z(v_i)-Z(\pi(v_i)),
\end{equation}
where $\pi(v_i)$ denotes the direct parent of $v_i$.
A positive $\Delta Z(v_i)$ indicates an increase in
the anomaly score relative to the parent.
For the root, we set $\Delta Z=0$.

We combine the node's normalized anomaly score with
its positive parent-relative increment to obtain
the final ranking score:
\begin{equation}
    R(v_i)=Z(v_i)+\alpha\max(0,\Delta Z(v_i)),
\end{equation}
where $\alpha \geq 0$ weights the positive parent-relative
increment, prioritizing nodes with high anomaly scores
or sharp increases over their parents.

\textbf{Context-Preserving Attribution.}
We select the top-$K$ nodes ranked by $R(v_i)$.
For each candidate, we retain its parent, the node
itself, and its direct children, then deduplicate
overlapping context.
Original contents, node identifiers, execution order,
and retained dependency edges preserve the upstream
premise and downstream consequences.
Together with the initial query, this evidence
is passed to an LLM-based attribution method,
such as CORRECT~\citep{yu2025correct} or
ECHO~\citep{banerjee2025echo}, to identify the
responsible agent and failure step.

\section{Experiments}
\definecolor{deltatext}{RGB}{178,34,34}
\definecolor{deltabg}{RGB}{253,240,240}

\subsection{Experimental Setup}
\label{sec:exp_setup}

\textbf{Dataset.}
We evaluate \textsc{DuoTrace} on the \textit{Who\&When} benchmark~\citep{zhang2025who}, which contains 184 failure instances. Its Handcraft subset comprises 58 instances with longer, more complex trajectories, while Automated contains 126 instances with generally shorter executions.
Additional task and failure-category statistics
are provided in Appendix~\ref{sec:appendix_data_stats}.

\textbf{Baselines.}
We evaluate \textsc{DuoTrace} using direct prompting with
DeepSeek-R1~\citep{deepseek2025r1},
Claude-Sonnet-4~\citep{anthropic2025claude4},
GPT-5~\citep{openai2025gpt5}, and
Qwen3.5-plus~\citep{qwen2025qwen35},
alongside two LLM-based attribution frameworks,
ECHO~\citep{banerjee2025echo} and
CORRECT~\citep{yu2025correct},
both using Qwen3.5-plus as their backbone.

\textbf{Evaluation Metrics.}
Agent and step accuracy measure attribution performance.
Overall accuracy is the weighted average of Handcraft and
Automated results, with weights $58/184$ and $126/184$.
We evaluate candidate selection using R@20\%, the fraction
of instances whose annotated root cause appears among the
top 20\% of ranked nodes.
Accuracy and recall differences are reported as absolute
changes in percentage points, while reductions in token
consumption and inference time are reported as relative
percentages. Model configurations and training details are
provided in Appendix~\ref{sec:appendix_implementation}.

\begin{table}[t]
  \centering
  \caption{Performance comparison of different methods. For each LLM-based method, we report the original result, its \textsc{DuoTrace}-enhanced counterpart, and the absolute improvement. }
  \label{tab:performance_comparison}
  \setlength{\tabcolsep}{2.5pt}
  \renewcommand{\arraystretch}{1.12}
  \resizebox{\columnwidth}{!}{
  \begin{tabular}{lllcccccc}
    \toprule
    \multirow{2}{*}{\textbf{Type}} 
    & \multirow{2}{*}{\textbf{Base Method}} 
    & \multirow{2}{*}{\textbf{Variant}} 
    & \multicolumn{2}{c}{\textbf{Handcraft}} 
    & \multicolumn{2}{c}{\textbf{Automated}} 
    & \multicolumn{2}{c}{\textbf{Overall}} \\
    \cmidrule(lr){4-5} \cmidrule(lr){6-7} \cmidrule(lr){8-9}
    & & & Agent & Step & Agent & Step & Agent & Step \\
    \midrule

    \multirow{2}{*}{Non-LLM}
    & FAMAS
    & -- 
    & 62.1 & \textbf{41.4} 
    & 55.6 & 23.8 
    & 57.6 & 29.3 \\
    & CDC-MAS 
    & -- 
    & 56.9 & 17.2 
    & 48.4 & 36.5 
    & 51.1 & 30.4 \\

    \midrule
\multirow{12}{*}{\begin{tabular}{@{}l@{}}LLM-based \\ Prompting\end{tabular}}
    & \multirow{3}{*}{DeepSeek-R1}
    & Original 
    & 53.4 & 6.9 
    & 57.1 & 29.4 
    & 55.9 & 22.3 \\
    & 
    & \cellcolor{blue!6}{+\textsc{DuoTrace}}
    & \cellcolor{blue!6}{62.1} & \cellcolor{blue!6}{13.8} 
    & \cellcolor{blue!6}{\underline{66.7}} & \cellcolor{blue!6}{35.7} 
    & \cellcolor{blue!6}{65.3} & \cellcolor{blue!6}{28.8} \\
    &
    & \cellcolor{green!8}{$\Delta$}
    & \cellcolor{green!8}{+8.7} & \cellcolor{green!8}{+6.9}
    & \cellcolor{green!8}{+9.6} & \cellcolor{green!8}{+6.3}
    & \cellcolor{green!8}{+9.4} & \cellcolor{green!8}{+6.5} \\

    \cmidrule(lr){2-9}

    & \multirow{3}{*}{\begin{tabular}{@{}l@{}}Claude-\\Sonnet-4\end{tabular}}
    & Original
    & 50.0 & 19.0
    & 50.8 & 38.9
    & 50.5 & 32.6 \\
    &
    & \cellcolor{blue!6}{+\textsc{DuoTrace}}
    & \cellcolor{blue!6}{60.3} & \cellcolor{blue!6}{24.1}
    & \cellcolor{blue!6}{57.1} & \cellcolor{blue!6}{42.9}
    & \cellcolor{blue!6}{58.1} & \cellcolor{blue!6}{37.0} \\
    &
    & \cellcolor{green!8}{$\Delta$}
    & \cellcolor{green!8}{+10.3} & \cellcolor{green!8}{+5.1}
    & \cellcolor{green!8}{+6.3} & \cellcolor{green!8}{+4.0}
    & \cellcolor{green!8}{+7.6} & \cellcolor{green!8}{+4.4} \\

    \cmidrule(lr){2-9}

    & \multirow{3}{*}{GPT-5}
    & Original 
    & 36.2 & 12.1 
    & 48.4 & 18.3 
    & 44.6 & 16.3 \\
    &
    & \cellcolor{blue!6}{+\textsc{DuoTrace}}
    & \cellcolor{blue!6}{58.6} & \cellcolor{blue!6}{22.4} 
    & \cellcolor{blue!6}{64.3} & \cellcolor{blue!6}{33.3} 
    & \cellcolor{blue!6}{62.5} & \cellcolor{blue!6}{29.9} \\
    &
    & \cellcolor{green!8}{$\Delta$}
    & \cellcolor{green!8}{+22.4} & \cellcolor{green!8}{+10.3}
    & \cellcolor{green!8}{+15.9} & \cellcolor{green!8}{+15.0}
    & \cellcolor{green!8}{+17.9} & \cellcolor{green!8}{+13.6} \\

\cmidrule(lr){2-9}
& \multirow{3}{*}{Qwen3.5-plus}
    & Original 
    & 53.4 & 17.2 
    & 54.0 & 38.1 
    & 53.8 & 31.5 \\
    &
    & \cellcolor{blue!6}{+\textsc{DuoTrace}}
    & \cellcolor{blue!6}{\textbf{74.1}} & \cellcolor{blue!6}{25.9} 
    & \cellcolor{blue!6}{61.9} & \cellcolor{blue!6}{46.0} 
    & \cellcolor{blue!6}{\underline{65.7}} & \cellcolor{blue!6}{39.7} \\
    &
    & \cellcolor{green!8}{$\Delta$}
    & \cellcolor{green!8}{+20.7} & \cellcolor{green!8}{+8.7}
    & \cellcolor{green!8}{+7.9} & \cellcolor{green!8}{+7.9}
    & \cellcolor{green!8}{+11.9} & \cellcolor{green!8}{+8.2} \\
    \midrule

    \multirow{7}{*}{\begin{tabular}{@{}l@{}}LLM-as- \\ a-Judge\end{tabular}}
    & AgenTracer
    & -- 
    & 63.8 & 20.7 
    & 63.5 & 37.3 
    & 63.6 & 32.1 \\

    \cmidrule(lr){2-9}

    & \multirow{3}{*}{ECHO}
    & Original 
    & 48.3 & 22.4 
    & 57.9 & 42.1 
    & 54.9 & 35.9 \\
    &
    & \cellcolor{blue!6}{+\textsc{DuoTrace}}
    & \cellcolor{blue!6}{63.8} & \cellcolor{blue!6}{25.9} 
    & \cellcolor{blue!6}{65.9} & \cellcolor{blue!6}{48.4} 
    & \cellcolor{blue!6}{65.2} & \cellcolor{blue!6}{41.3} \\
    &
    & \cellcolor{green!8}{$\Delta$}
    & \cellcolor{green!8}{+15.5} & \cellcolor{green!8}{+3.5}
    & \cellcolor{green!8}{+8.0} & \cellcolor{green!8}{+6.3}
    & \cellcolor{green!8}{+10.3} & \cellcolor{green!8}{+5.4} \\

    \cmidrule(lr){2-9}

    & \multirow{3}{*}{CORRECT}
    & Original 
    & 58.6 & 20.7 
    & 64.3 & \underline{51.6} 
    & 62.5 & \underline{41.9} \\
    &
    & \cellcolor{blue!6}{+\textsc{DuoTrace}}
    & \cellcolor{blue!6}{\underline{72.4}} & \cellcolor{blue!6}{\underline{31.0}} 
    & \cellcolor{blue!6}{\textbf{72.2}} & \cellcolor{blue!6}{\textbf{58.7}} 
    & \cellcolor{blue!6}{\textbf{72.3}} & \cellcolor{blue!6}{\textbf{50.0}} \\
    &
    & \cellcolor{green!8}{$\Delta$}
    & \cellcolor{green!8}{+13.8} & \cellcolor{green!8}{+10.3}
    & \cellcolor{green!8}{+7.9} & \cellcolor{green!8}{+7.1}
    & \cellcolor{green!8}{+9.8} & \cellcolor{green!8}{+8.1} \\

    \bottomrule
  \end{tabular}
  }
\end{table}

\subsection{Overall Performance}
\label{sec:overall_performance}
\textbf{State-of-the-Art Accuracy and Fine-Grained Diagnostics.}
Table~\ref{tab:performance_comparison} shows that CORRECT
with \textsc{DuoTrace} achieves the highest overall agent
and step accuracy of 72.3\% and 50.0\%, respectively,
with an absolute 8.1\% gain in step accuracy over
unenhanced CORRECT.
These improvements suggest that effective LLM-based
attribution benefits from focused evidence that limits
irrelevant execution details while preserving the
invocation context needed to distinguish initiating
errors from downstream consequences.

\textbf{Consistent Plug-and-Play Improvements.}
\textsc{DuoTrace} improves agent and step accuracy across
all six downstream methods on both subsets, covering
direct prompting and specialized attribution frameworks.
GPT-5 benefits most, with gains of 17.9\% and 13.6\%
in overall agent and step accuracy, respectively.
The gains with ECHO and CORRECT further demonstrate that
candidate filtering complements structured attribution procedures.

% % === 第一张图：Agent-level 增量图 ===
% \begin{figure}[htbp]
%     \centering
%     \includegraphics[width=\columnwidth]{agent_accuracy_HC_new.pdf}% 您的 Agent-level 图片文件名
%     \caption{Absolute Agent-level performance improvement (\%) across different foundation models and specialized frameworks on the Handcraft dataset.}
%     \label{fig:agent_accuracy_hc}
% \end{figure}

% % === 第二张图：Step-level 增量图 ===
% \begin{figure}[htbp]
%     \centering
%     \includegraphics[width=\columnwidth]{step_accuracy_HC_new.pdf} % 请替换为您的 Step-level 图片文件名
%     \caption{Absolute Step-level performance improvement (\%) across different foundation models and specialized frameworks on the Handcraft dataset.}
%     \label{fig:step_accuracy_hc}
% \end{figure}

\subsection{Ablation Study}
\label{sec:ablation_study}

\textbf{Data Augmentation.}
Removing outcome-verified replay or root-to-node
path extraction lowers recall on both subsets, with larger declines on Handcraft
(Table~\ref{tab:ablation_data_aug}).
Replay removal has the greater impact, supporting the
value of diverse successful executions alongside
intermediate-state coverage.

\textbf{Model Components.}
Removing the semantic or structural view produces
the largest recall drops (Table~\ref{tab:ablation_vae}),
supporting their complementary contributions.
Without contextual semantic reconstruction, recall
falls to 74.14\% on Handcraft and 92.86\% on Automated.
Ablating agent, operator, token-prediction, or KL losses
also lowers recall on both subsets, highlighting
the value of the combined training objective.
\begin{figure}[t]
  \centering
  \includegraphics[width=\linewidth]{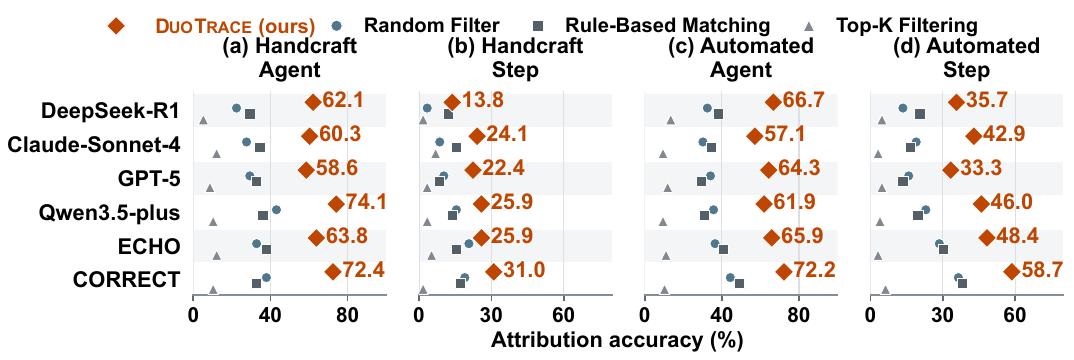}
  \caption{Candidate filtering comparison.}
  \label{fig:filtering_comparison}
\end{figure}

\textbf{Candidate Filtering.}
We compare \textsc{DuoTrace} with three baselines,
each selecting the same number of candidate nodes.
Random Filter samples nodes uniformly at random;
Rule-Based Matching selects nodes containing failure
keywords (e.g., ``error'', ``failed'', ``invalid'');
and Top-K Filtering selects the first $K$ nodes
in execution order.
Figure~\ref{fig:filtering_comparison} shows that
\textsc{DuoTrace} outperforms all three baselines
across all evaluated methods and metrics.

\begin{table}[t]
  \centering
  \caption{Ablation study on data augmentation and VAE components.
  R@20\% denotes root-cause recall (\%) among the top 20\%
  of ranked nodes. $\Delta$ denotes the change from the
  full model in percentage points.}
  \label{tab:ablation_combined}
  \label{tab:ablation_data_aug}
  \label{tab:ablation_vae}

  % Keep the font size consistent with the other tables.
  \fontsize{8pt}{9.6pt}\selectfont
  \setlength{\tabcolsep}{2pt}
  \setlength{\fboxsep}{1pt}
  \renewcommand{\arraystretch}{1.15}

  % Keep variant names on one line.
  % The two recall columns share the remaining width.
  % Reduce the gap within each recall--delta pair to 1pt.
  \begin{tabularx}{\columnwidth}{
    @{}
    l
    @{\hspace{3pt}}
    >{\centering\arraybackslash}X
    @{\hspace{1pt}}
    c
    @{\hspace{4pt}}
    >{\centering\arraybackslash}X
    @{\hspace{1pt}}
    c
    @{}
  }
    \toprule
    \multirow{2}{*}{\textbf{Variant}}
    & \multicolumn{2}{c}{\textbf{Handcraft}}
    & \multicolumn{2}{c}{\textbf{Automated}} \\
    \cmidrule(lr){2-3}
    \cmidrule(lr){4-5}
    & \textbf{R@20\%} & $\boldsymbol{\Delta}$
    & \textbf{R@20\%} & $\boldsymbol{\Delta}$ \\
    \midrule

    \textbf{Full Model}
    & \textbf{91.38} & --
    & \textbf{98.41} & -- \\
    \midrule

    \multicolumn{5}{@{}l@{}}{\textbf{Data augmentation}} \\
    \addlinespace[2pt]

    \textit{w/o} Root-to-Node Path Extraction
    & 72.41
    & \colorbox{deltabg}{\textcolor{deltatext}{$-18.97$}}
    & 92.86
    & \colorbox{deltabg}{\textcolor{deltatext}{$-5.55$}} \\

    \textit{w/o} Outcome-Verified Replay
    & 70.69
    & \colorbox{deltabg}{\textcolor{deltatext}{$-20.69$}}
    & 90.48
    & \colorbox{deltabg}{\textcolor{deltatext}{$-7.93$}} \\
    \midrule

    \multicolumn{5}{@{}l@{}}{\textbf{VAE components}} \\
    \addlinespace[2pt]

    \textit{w/o} Structure View
    & 70.69
    & \colorbox{deltabg}{\textcolor{deltatext}{$-20.69$}}
    & 83.33
    & \colorbox{deltabg}{\textcolor{deltatext}{$-15.08$}} \\

    \textit{w/o} Semantic View
    & 60.34
    & \colorbox{deltabg}{\textcolor{deltatext}{$-31.04$}}
    & 82.54
    & \colorbox{deltabg}{\textcolor{deltatext}{$-15.87$}} \\

    \textit{w/o} Agent Loss
    & 74.14
    & \colorbox{deltabg}{\textcolor{deltatext}{$-17.24$}}
    & 92.06
    & \colorbox{deltabg}{\textcolor{deltatext}{$-6.35$}} \\

    \textit{w/o} Operator Loss
    & 72.41
    & \colorbox{deltabg}{\textcolor{deltatext}{$-18.97$}}
    & 85.71
    & \colorbox{deltabg}{\textcolor{deltatext}{$-12.70$}} \\

    \textit{w/o} Contextual Semantic Loss
    & 74.14
    & \colorbox{deltabg}{\textcolor{deltatext}{$-17.24$}}
    & 92.86
    & \colorbox{deltabg}{\textcolor{deltatext}{$-5.55$}} \\

    \textit{w/o} NLL Loss
    & 77.59
    & \colorbox{deltabg}{\textcolor{deltatext}{$-13.79$}}
    & 87.30
    & \colorbox{deltabg}{\textcolor{deltatext}{$-11.11$}} \\

    \textit{w/o} KL Diverg. (AE)
    & 81.03
    & \colorbox{deltabg}{\textcolor{deltatext}{$-10.35$}}
    & 90.48
    & \colorbox{deltabg}{\textcolor{deltatext}{$-7.93$}} \\

    \bottomrule
  \end{tabularx}
\end{table}

\subsection{Computational Overhead and Practicality}
\label{sec:discussion_efficiency}

\begin{figure}[t]
  \centering
  
  % (a) 左侧子图：Token 消耗
  \begin{subfigure}{0.49\columnwidth} % �� 核心修改2：改为 \columnwidth
    \centering
    \includegraphics[width=\linewidth]{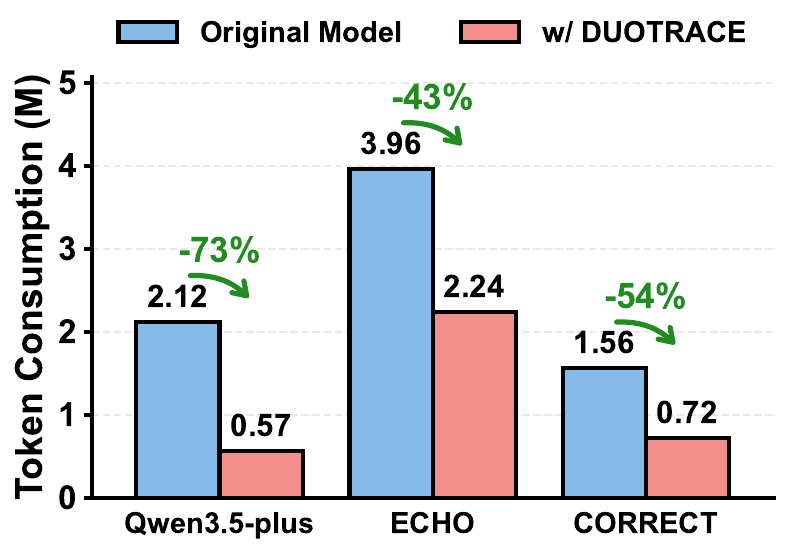} 
    \caption{Token Consumption}
    \label{fig:efficiency_tokens}
  \end{subfigure}% <-- 加上百分号防止多余的换行空格
  \hfill
  % (b) 右侧子图：推理时间
  \begin{subfigure}{0.49\columnwidth} % �� 核心修改2：改为 \columnwidth
    \centering
    % 注意：刚才 Python 代码里生成的名字是 inference_time_HC.pdf
    \includegraphics[width=\linewidth]{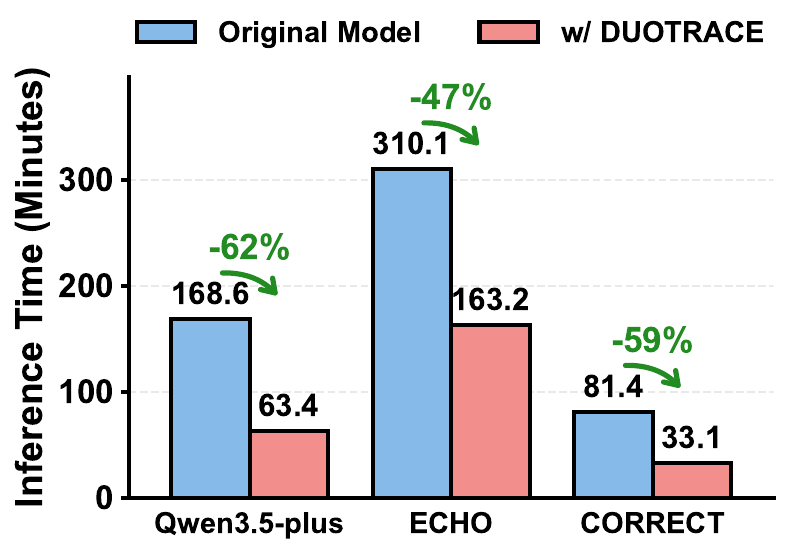} 
    \caption{Inference Time}
    \label{fig:efficiency_time}
  \end{subfigure}
  
\caption{Comparison of computational overhead and inference latency on the Handcraft dataset for Qwen3.5-plus, ECHO, and CORRECT. The \textsc{DuoTrace} filter reduces both token consumption (a) and inference time (b) for these methods.}

\label{fig:efficiency_comparison}
\end{figure}

Figure~\ref{fig:efficiency_comparison} compares token consumption and inference time on Handcraft. The reported reductions vary by downstream method: Qwen3.5-plus uses approximately 73\% fewer tokens, while ECHO uses approximately 43\% fewer. CORRECT's inference time decreases from 81.4 to 33.1 minutes, corresponding to a 2.46$\times$ speedup. Together with the accuracy improvements in Table~\ref{tab:performance_comparison}, these results support concentrating downstream computation on selected trajectory evidence.

% \subsection{RQ3: Counterfactual Alignment Analysis}
% \label{sec:rq3_alignment}

% To validate our data augmentation, we apply Dynamic Time Warping (DTW) \citep{muller2007dynamic} to align generated successful trajectories with original failed logs. Figure \ref{fig:alignment_trend} illustrates the semantic similarity trend measured via BERTScore \citep{zhang2019bertscore}.

% \textbf{Prefix Consistency ($X < 0$).} Similarity remains high ($>0.83$, peaking at 0.96) before the failure point, confirming that the augmentation preserves the original operational preconditions and context.

% \textbf{Divergence at the Failure Node ($X = 0$).} At the exact failure step, similarity drops sharply (from 0.72 to 0.44 at $X=1$). This discontinuity proves the generative model accurately targets and corrects the specific root cause, rather than drifting randomly.

% \textbf{Post-Correction Diversity ($X > 0$).} After intervention, similarity stabilizes at a lower baseline ($\sim$0.34--0.39). This deliberate semantic distinctiveness provides the VAE with diverse positive samples, enriching its capacity to model valid execution variations.

\begin{table}[t]
  \centering
  \caption{Comparison of encoder architectures under identical experimental
settings on the Handcraft and Automated subsets.}
  \label{tab:discussion_encoder}

  % Identical font size and spacing in all four tables.
  \fontsize{8pt}{9.6pt}\selectfont
  \setlength{\tabcolsep}{2pt}
  \renewcommand{\arraystretch}{1.15}

  % X column adapts to the available single-column width.
  \begin{tabularx}{\columnwidth}{
    @{} l >{\raggedright\arraybackslash}X c c c c @{}
  }
    \toprule
    \multirow{2}{*}{\textbf{Method}}
    & \multirow{2}{*}{\textbf{Encoder}}
    & \multicolumn{2}{c}{\textbf{Handcraft}}
    & \multicolumn{2}{c}{\textbf{Automated}} \\
    \cmidrule(lr){3-4}
    \cmidrule(lr){5-6}
    & & \textbf{Agent} & \textbf{Step}
      & \textbf{Agent} & \textbf{Step} \\
    \midrule

    \multirow{5}{*}{Qwen3.5-plus}
    & MLP
    & 48.3 & 12.1 & 40.5 & 29.4 \\

    & Transformer
    & 53.4 & 17.2 & 45.2 & 31.7 \\

    & Tree-Transformer
    & 67.2 & 22.4 & 55.6 & 42.1 \\

    & GAT
    & 65.5 & 20.7 & 51.6 & 39.7 \\

    & \textbf{Tree-LSTM}
    & \textbf{74.1} & \textbf{25.9}
    & \textbf{61.9} & \textbf{46.0} \\
    \midrule

    \multirow{5}{*}{CORRECT}
    & MLP
    & 51.7 & 10.3 & 46.0 & 32.5 \\

    & Transformer
    & 56.9 & 22.4 & 54.0 & 40.5 \\

    & Tree-Transformer
    & 67.2 & 29.3 & 63.5 & 53.2 \\

    & GAT
    & 62.1 & 27.6 & 60.3 & 49.2 \\

    & \textbf{Tree-LSTM}
    & \textbf{72.4} & \textbf{31.0}
    & \textbf{72.2} & \textbf{58.7} \\

    \bottomrule
  \end{tabularx}
\end{table}

\subsection{Encoder Architecture Analysis}

Table~\ref{tab:discussion_encoder} compares five encoders under matched settings, using Qwen3.5-plus and CORRECT as downstream methods. The sequential Transformer consistently improves over the MLP, and all three structure-aware encoders outperform both alternatives. This ordering supports encoding invocation dependencies rather than relying exclusively on flat or sequential representations.

Tree-LSTM achieves the best agent and step accuracy across
all evaluated settings. With CORRECT, it outperforms
Tree-Transformer, the strongest alternative, by 8.7\% and
5.5\% in Automated agent and step accuracy, respectively.
These results support hierarchical recurrent aggregation
for the evaluated execution traces.
\begin{table}[t]
  \centering

\caption{Robustness to random masking of 20\% of semantic fields with execution topology preserved. Agent and step accuracy (\%) are reported on Handcraft and Automated; $\Delta$ denotes the change in percentage points relative to the corresponding unmasked setting.}

  \label{tab:robustness_masking}

  \resizebox{\columnwidth}{!}{
  \begin{tabular}{llcccccccc}
    \toprule
    \multirow{2}{*}{\textbf{Method}}
    & \multirow{2}{*}{\textbf{Variant}}
    & \multicolumn{4}{c}{\textbf{Handcraft}}
    & \multicolumn{4}{c}{\textbf{Automated}} \\
    \cmidrule(lr){3-6}
    \cmidrule(lr){7-10}
    &
    & \textbf{Agent} & $\boldsymbol{\Delta}$
    & \textbf{Step}  & $\boldsymbol{\Delta}$
    & \textbf{Agent} & $\boldsymbol{\Delta}$
    & \textbf{Step}  & $\boldsymbol{\Delta}$ \\
    \midrule

    % GPT-5
    \multirow{4}{*}{GPT-5}
    & Baseline
    & 36.2 & -- & 12.1 & -- & 48.4 & -- & 18.3 & -- \\

    & \cellcolor{blue!6}{Baseline + Mask}
    & \cellcolor{blue!6}{20.7} & \cellcolor{green!8}{-15.5}
    & \cellcolor{blue!6}{6.9}  & \cellcolor{green!8}{-5.2}
    & \cellcolor{blue!6}{39.7} & \cellcolor{green!8}{-8.7}
    & \cellcolor{blue!6}{13.5} & \cellcolor{green!8}{-4.8} \\

    & \textsc{DuoTrace}
    & 58.6 & -- & 22.4 & -- & 64.3 & -- & 33.3 & -- \\

    & \cellcolor{blue!6}{\textsc{DuoTrace} + Mask}
    & \cellcolor{blue!6}{50.0} & \cellcolor{green!8}{-8.6}
    & \cellcolor{blue!6}{19.0} & \cellcolor{green!8}{-3.4}
    & \cellcolor{blue!6}{59.5} & \cellcolor{green!8}{-4.8}
    & \cellcolor{blue!6}{29.4} & \cellcolor{green!8}{-3.9} \\
    \midrule

    % Qwen3.5-plus
    \multirow{4}{*}{Qwen3.5-plus}
    & Baseline
    & 53.4 & -- & 17.2 & -- & 54.0 & -- & 38.1 & -- \\

    & \cellcolor{blue!6}{Baseline + Mask}
    & \cellcolor{blue!6}{36.2} & \cellcolor{green!8}{-17.2}
    & \cellcolor{blue!6}{10.3} & \cellcolor{green!8}{-6.9}
    & \cellcolor{blue!6}{44.4} & \cellcolor{green!8}{-9.6}
    & \cellcolor{blue!6}{31.7} & \cellcolor{green!8}{-6.4} \\

    & \textsc{DuoTrace}
    & 74.1 & -- & 25.9 & -- & 61.9 & -- & 46.0 & -- \\

    & \cellcolor{blue!6}{\textsc{DuoTrace} + Mask}
    & \cellcolor{blue!6}{63.8} & \cellcolor{green!8}{-10.3}
    & \cellcolor{blue!6}{22.4} & \cellcolor{green!8}{-3.5}
    & \cellcolor{blue!6}{57.1} & \cellcolor{green!8}{-4.8}
    & \cellcolor{blue!6}{42.1} & \cellcolor{green!8}{-3.9} \\
    \midrule

    % ECHO
    \multirow{4}{*}{ECHO}
    & Baseline
    & 48.3 & -- & 22.4 & -- & 57.9 & -- & 42.1 & -- \\

    & \cellcolor{blue!6}{Baseline + Mask}
    & \cellcolor{blue!6}{37.9} & \cellcolor{green!8}{-10.4}
    & \cellcolor{blue!6}{15.5} & \cellcolor{green!8}{-6.9}
    & \cellcolor{blue!6}{53.2} & \cellcolor{green!8}{-4.7}
    & \cellcolor{blue!6}{34.1} & \cellcolor{green!8}{-8.0} \\

    & \textsc{DuoTrace}
    & 63.8 & -- & 25.9 & -- & 65.9 & -- & 48.4 & -- \\

    & \cellcolor{blue!6}{\textsc{DuoTrace} + Mask}
    & \cellcolor{blue!6}{56.9} & \cellcolor{green!8}{-6.9}
    & \cellcolor{blue!6}{24.1} & \cellcolor{green!8}{-1.8}
    & \cellcolor{blue!6}{63.5} & \cellcolor{green!8}{-2.4}
    & \cellcolor{blue!6}{44.4} & \cellcolor{green!8}{-4.0} \\
    \midrule

    % CORRECT
    \multirow{4}{*}{CORRECT}
    & Baseline
    & 58.6 & -- & 20.7 & -- & 64.3 & -- & 51.6 & -- \\

    & \cellcolor{blue!6}{Baseline + Mask}
    & \cellcolor{blue!6}{46.6} & \cellcolor{green!8}{-12.0}
    & \cellcolor{blue!6}{15.5} & \cellcolor{green!8}{-5.2}
    & \cellcolor{blue!6}{55.6} & \cellcolor{green!8}{-8.7}
    & \cellcolor{blue!6}{43.7} & \cellcolor{green!8}{-7.9} \\

    & \textsc{DuoTrace}
    & 72.4 & -- & 31.0 & -- & 72.2 & -- & 58.7 & -- \\

    & \cellcolor{blue!6}{\textsc{DuoTrace} + Mask}
    & \cellcolor{blue!6}{67.2} & \cellcolor{green!8}{-5.2}
    & \cellcolor{blue!6}{29.3} & \cellcolor{green!8}{-1.7}
    & \cellcolor{blue!6}{66.7} & \cellcolor{green!8}{-5.5}
    & \cellcolor{blue!6}{54.8} & \cellcolor{green!8}{-3.9} \\

    \bottomrule
  \end{tabular}
  }
\end{table}

\subsection{Robustness Analysis}
\label{sec:robustness}

\textbf{Partial Semantic Observability.}
With 20\% of semantic fields randomly masked and topology
preserved, \textsc{DuoTrace} reduces the mean accuracy drop
from 8.6\% to 4.7\% across 16 settings
(Table~\ref{tab:robustness_masking}).
All masked \textsc{DuoTrace} variants outperform their
unmasked baselines, supporting robustness to incomplete
semantic information.

\textbf{Non-Tree Execution Structures.}
We evaluate non-tree traces from LongRCA~\citep{zhang2026longrca}
using structural unrolling to obtain tree-compatible
representations. \textsc{DuoTrace} improves both attribution
metrics across all four downstream methods, spanning direct
prompting and specialized attribution frameworks
(Table~\ref{tab:non_tree_robustness}). CORRECT enhanced with
\textsc{DuoTrace} achieves the best results on both metrics,
supporting the use of candidate filtering on non-tree traces
after structural unrolling.
\begin{table}[t]
  \centering
  \caption{Attribution performance on non-tree traces from LongRCA after structural unrolling.}
\label{tab:non_tree_robustness}
  \fontsize{8pt}{9.6pt}\selectfont
  \setlength{\tabcolsep}{2pt}
  \renewcommand{\arraystretch}{1.15}

  \begin{tabularx}{\columnwidth}{
    @{} l >{\raggedright\arraybackslash}X c c @{}
  }
    \toprule
    \textbf{Type}
    & \textbf{Method}
    & \textbf{Agent}
    & \textbf{Step} \\
    \midrule

    \multirow{4}{*}{\shortstack[l]{LLM-based\\Prompting}}
    & GPT-5
    & 38.1 & 28.6 \\

    & \textbf{GPT-5 + \textsc{DuoTrace}}
    & \textbf{57.1} & \textbf{42.9} \\

    & Qwen3.5-plus
    & 33.3 & 19.0 \\

    & \textbf{Qwen3.5-plus + \textsc{DuoTrace}}
    & \textbf{47.6} & \textbf{28.6} \\
    \midrule

    \multirow{4}{*}{\shortstack[l]{LLM-as-\\a-Judge}}
    & ECHO
    & 57.1 & 42.9 \\

    & \textbf{ECHO + \textsc{DuoTrace}}
    & \textbf{66.7} & \textbf{52.4} \\

    & CORRECT
    & 61.9 & 52.4 \\

    & \textbf{CORRECT + \textsc{DuoTrace}}
    & \textbf{76.2} & \textbf{61.9} \\

    \bottomrule
  \end{tabularx}
\end{table}

\begin{table}[t]
  \centering
\caption{Cross-benchmark generalization under single-root-cause
settings on AgentErrorBench (AEBench) and RealWorld. Ag and St denote agent and step accuracy (\%),
respectively.}
  \label{tab:cross_benchmark_single}

  \fontsize{8pt}{9.6pt}\selectfont
  \setlength{\tabcolsep}{2pt}
  \renewcommand{\arraystretch}{1.15}

  \begin{tabularx}{\columnwidth}{
    @{} l >{\raggedright\arraybackslash}X c c c c @{}
  }
    \toprule
    \multirow{2}{*}{\textbf{Type}}
    & \multirow{2}{*}{\textbf{Method}}
    & \multicolumn{2}{c}{\textbf{AEBench}}
    & \multicolumn{2}{c}{\textbf{RealWorld}} \\
    \cmidrule(lr){3-4}
    \cmidrule(lr){5-6}
    & & \textbf{Ag} & \textbf{St}
      & \textbf{Ag} & \textbf{St} \\
    \midrule

    \multirow{4}{*}{\shortstack[l]{LLM-based\\Prompting}}
    & GPT-5
    & 37.9 & 14.6 & 51.0 & 5.2 \\

    & \textbf{GPT-5 + \textsc{DuoTrace}}
    & \textbf{47.6} & \textbf{23.3}
    & \textbf{62.5} & \textbf{14.6} \\

    & Qwen3.5-plus
    & 41.7 & 15.5 & 54.2 & 7.3 \\

    & \textbf{Qwen3.5-plus + \textsc{DuoTrace}}
    & \textbf{52.4} & \textbf{24.2}
    & \textbf{65.6} & \textbf{16.7} \\
    \midrule

    \multirow{4}{*}{\shortstack[l]{LLM-as-\\a-Judge}}
    & ECHO
    & 47.5 & 18.4 & 53.1 & 4.2 \\

    & \textbf{ECHO + \textsc{DuoTrace}}
    & \textbf{59.7} & \textbf{27.7}
    & \textbf{66.7} & \textbf{17.7} \\

    & CORRECT
    & 50.5 & 21.4 & 56.3 & 9.4 \\

    & \textbf{CORRECT + \textsc{DuoTrace}}
    & \textbf{63.6} & \textbf{30.1}
    & \textbf{68.8} & \textbf{22.9} \\

    \bottomrule
  \end{tabularx}
\end{table}

\begin{table}[t]
  \centering
  \caption{Candidate-ranking performance on MP-Bench under
  multi-cause settings. Higher nDCG@5 indicates better
  ranking quality.}
  \label{tab:mpbench_multicause}

  \fontsize{8pt}{9.6pt}\selectfont
  \setlength{\tabcolsep}{2pt}
  \renewcommand{\arraystretch}{1.15}

  \begin{tabularx}{\columnwidth}{
    @{} l >{\raggedright\arraybackslash}X c c @{}
  }
    \toprule
    \textbf{Type}
    & \textbf{Method}
    & \shortstack{\textbf{Manual}\\\textbf{nDCG@5}}
    & \shortstack{\textbf{Auto}\\\textbf{nDCG@5}} \\
    \midrule

    \multirow{4}{*}{\shortstack[l]{LLM-based\\Prompting}}
    & GPT-5
    & 0.3643 & 0.6432 \\

    & \textbf{GPT-5 + \textsc{DuoTrace}}
    & \textbf{0.4382} & \textbf{0.7153} \\

    & Qwen3.5-plus
    & 0.3824 & 0.7021 \\

    & \textbf{Qwen3.5-plus + \textsc{DuoTrace}}
    & \textbf{0.4417} & \textbf{0.7563} \\
    \midrule

    \multirow{4}{*}{\shortstack[l]{LLM-as-\\a-Judge}}
    & ECHO
    & 0.4243 & 0.7122 \\

    & \textbf{ECHO + \textsc{DuoTrace}}
    & \textbf{0.4582} & \textbf{0.7741} \\

    & CORRECT
    & 0.4157 & 0.7283 \\

    & \textbf{CORRECT + \textsc{DuoTrace}}
    & \textbf{0.4662} & \textbf{0.7892} \\

    \bottomrule
  \end{tabularx}
\end{table}

\subsection{Generalization Across Benchmarks and Failure Settings}
\label{sec:generalization}
\textbf{Cross-Benchmark Attribution.}
We evaluate single-root-cause attribution on
AgentErrorBench~\citep{zhu2025where} and the RealWorld
subset of LongRCA Bench~\citep{zhang2026longrca}, whose
target schemas are unseen during training.
\textsc{DuoTrace} improves both metrics for every downstream
method on both benchmarks
(Table~\ref{tab:cross_benchmark_single}).
On RealWorld, step accuracy rises from 4.2\% to 17.7\%
for ECHO and from 9.4\% to 22.9\% for CORRECT.
Enhanced CORRECT leads both metrics on both benchmarks,
supporting the transfer of candidate filtering across
the evaluated execution schemas. Additional experiments on generalization to held-out
task categories are reported in Appendix~\ref{app:heldout_task}.

\textbf{Multi-Cause Candidate Ranking.}
MP-Bench~\citep{in2026mpbench} evaluates trajectories with multiple defensible failure causes. We measure candidate-ranking quality using nDCG@5. As shown in Table~\ref{tab:mpbench_multicause}, \textsc{DuoTrace} improves all four methods in both Manual and Auto settings. It also changes the leading method in Manual from ECHO to CORRECT, while enhanced CORRECT achieves the highest ranking quality in both settings. The improvements show that candidate filtering benefits the ordering of multiple failure-relevant nodes as well as single-root-cause attribution.

\begin{figure}[t]
    \centering
    % === 左图：Automated 数据集 ===
    \begin{subfigure}{0.49\columnwidth}
        \centering
        % width=\linewidth 确保图片撑满这个子区域，不会越界
        \includegraphics[width=\linewidth]{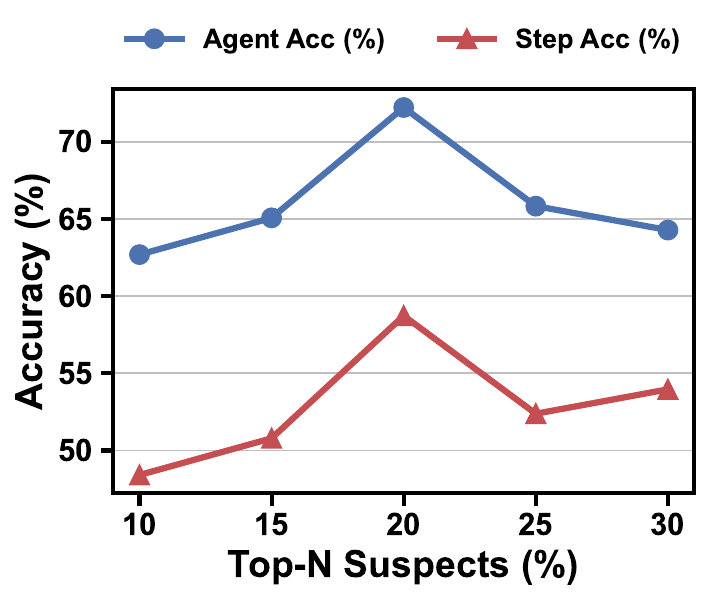}
        \caption{Automated}
        \label{fig:recall_ag}
    \end{subfigure}\hfill % \hfill 自动填充中间的空白，把两张图推向两边
    % === 右图：Handcraft 数据集 ===
    \begin{subfigure}{0.49\columnwidth}
        \centering
        \includegraphics[width=\linewidth]{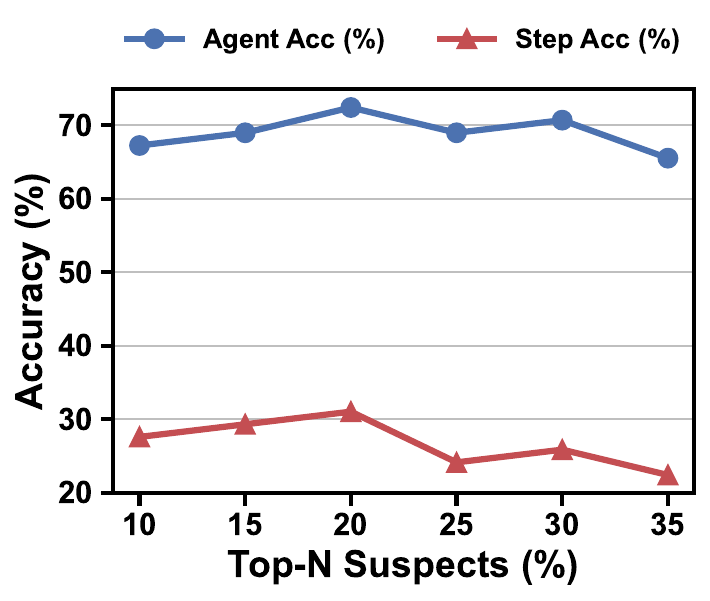}
        \caption{Handcraft}
        \label{fig:recall_hc}
    \end{subfigure}
    
    % \vspace{-2mm} % 如果觉得子图标题和总标题离得太远，可以取消这行的注释来微调间距
    \caption{Agent and step accuracy of CORRECT with
\textsc{DuoTrace} at different candidate proportions on Handcraft
and Automated. Proportions refer to selected candidate nodes
before local context expansion.}
    \label{fig:top_n_analysis}
\end{figure}
\subsection{Sensitivity to Candidate Set Proportion}

Figure~\ref{fig:top_n_analysis} shows that \textsc{DuoTrace}
maintains strong attribution performance across the tested
candidate proportions, indicating that its effectiveness
is not confined to a single optimal setting.
Both agent and step accuracy peak at 20\% on Handcraft
and Automated.
This proportion appears to balance evidence coverage
and context selectivity: smaller budgets may omit useful
evidence, while larger budgets introduce lower-ranked
nodes that can distract the judge.
We therefore retain the top 20\% of nodes as candidates.

\section{Conclusion}
% In this paper, we introduce \textsc{DuoTrace}, a Variational AutoEncoder (VAE)-enhanced framework designed to address the prohibitive token costs and execution uncertainties inherent in multi-agent failure attribution. By projecting noisy trajectory logs into a continuous probabilistic latent space, \textsc{DuoTrace} functions as a robust semantic filter. Comprehensive evaluations on the \textit{Who and When} dataset demonstrate that our plug-and-play module significantly amplifies the diagnostic capabilities of both foundation models and existing evaluators. Notably, the \textsc{DuoTrace}-CORRECT variant achieves an Agent Accuracy of 69.1\% and a Step Accuracy of 49.0\%. Furthermore, by effectively pruning redundant context, our framework reduces token consumption by approximately 41\% and accelerates inference by 2.5$\times$, facilitating scalable, real-world LLM-based debugging for complex multi-agent systems.
We propose \textsc{DuoTrace}, a plug-and-play
detect-before-attribute filter that learns normal
execution patterns from augmented successful trajectories.
Its dual-view Tree-LSTM VAE combines textual semantics
with invocation structure, using reconstruction errors
and parent-relative score changes to prioritize suspicious
nodes. The selected nodes and their local context provide
focused evidence for downstream LLM judges.
Experiments across six attribution methods demonstrate
consistent improvements in agent and step accuracy,
alongside reduced token consumption and inference time
in the evaluated efficiency settings.

\section*{Limitations}
\textsc{DuoTrace} relies on the quality and coverage of normal training trajectories. Outcome verification establishes task success but does not guarantee that every intermediate step is error-free, and generated traces may underrepresent real execution patterns. The Tree-LSTM encoder requires tree-compatible inputs; unrolling non-tree traces can duplicate shared nodes and alter how dependencies are represented. Finally, reconstruction errors and parent-relative changes measure deviations from learned execution patterns rather than causal responsibility. Benign deviations may receive high scores, while failures resembling normal behavior may be omitted from the retained evidence, limiting downstream attribution.

\section*{Ethics Statement}

Our research on \textsc{DuoTrace} aims to enhance the transparency, reliability, and safety of LLM-based multi-agent systems by providing accurate failure attribution. By effectively identifying the root causes of anomalous executions, we empower developers to build more robust AI agents. However, we acknowledge the potential dual-use nature of this technology. Advanced diagnostic frameworks that can precisely isolate system vulnerabilities could theoretically be exploited by malicious actors to craft targeted adversarial attacks or prompt injections against deployed agentic systems. We strongly advocate for the responsible use of such debugging tools strictly within secure, authorized development environments.

Regarding computational and environmental impacts, \textsc{DuoTrace} introduces a practical trade-off. While our ``detect-before-attribute'' framework significantly reduces inference overhead and token consumption during the downstream attribution phase---thereby promoting greener and more cost-effective LLM evaluation---the upstream process of training the Tree-LSTM-based VAE requires initial computational resources. We encourage future work to explore even lighter-weight architectures to further minimize the carbon footprint associated with AI evaluation pipelines.

Finally, our failure attribution process fundamentally relies on downstream LLMs acting as judges. These foundational models may possess intrinsic societal or cognitive biases, which could subtly influence the attribution outcomes and lead to skewed diagnostic results under specific contexts. We emphasize that all datasets and benchmarks used in our experiments are derived from simulated environments and standard open-source logs, containing no Personally Identifiable Information (PII) or offensive content.

We used AI assistants (e.g., ChatGPT) for LaTeX formatting, TikZ figure generation, and reference checking. All AI-generated content was reviewed and verified by the authors.

% 告诉 LaTeX 你的参考文献数据放在哪个文件里（不需要加 .bib 后缀）
\bibliography{custom}
% 在参考文献之后开始附录
\clearpage
\appendix
% ---------------------------------------------------------
% Section B: Dataset Statistics
% ---------------------------------------------------------
\section{Dataset Statistics}
\label{sec:appendix_data_stats}

% \subsection{Trajectory Length Distribution}
% Table~\ref{tab:step_dist} illustrate the distribution of the total number of steps per trajectory in the \textit{Who and When} benchmark. The dataset exhibits a highly skewed, long-tail distribution. Approximately 73.2\% of the trajectories are concise (5--20 steps), while the presence of a long-tail (traces extending up to 130 steps) highlights the substantial complexity inherent in multi-agent debugging. This justifies \textsc{DuoTrace}'s capability to distill long traces, effectively mitigating context dilution.

% \begin{table}[t]
%     \centering
%     \small
%     \caption{Distribution of trajectory lengths (Total Steps) in the benchmark.}
%     \label{tab:step_dist}
%     \begin{tabular}{lcc}
%         \toprule
%         \textbf{Step Range} & \textbf{Frequency} & \textbf{Percentage} \\
%         \midrule
%         5--20   & 142 & 77.2\% \\
%         21--40  & 16  & 8.7\% \\
%         41--70  & 10  & 5.4\% \\
%         71--100 & 6   & 3.3\% \\
%         101--130 & 10  & 5.4\% \\
%         \bottomrule
%     \end{tabular}
% \end{table}

\subsection{Task Classification Statistics}

Table~\ref{tab:task_taxonomy} summarizes the task
distribution across 184 failure instances.
Information Retrieval accounts for 72.3\% of the dataset,
while the remaining instances cover mathematical reasoning,
data analysis, multimodal processing, programming and
debugging, and text processing.
The dataset therefore spans several task categories,
with a pronounced concentration in information retrieval.

\begin{table}[t]
    \centering
    \caption{Task type distribution within the dataset.}
    \label{tab:task_taxonomy}

    \fontsize{8pt}{9.6pt}\selectfont
    \setlength{\tabcolsep}{2pt}
    \renewcommand{\arraystretch}{1.15}

    \begin{tabular*}{\columnwidth}{
        @{\extracolsep{\fill}} l c c @{}
    }
        \toprule
        \textbf{Task Category}
        & \textbf{Count}
        & \textbf{Percentage} \\
        \midrule

        Information Retrieval
        & 133 & 72.3\% \\

        Math \& Logic Reasoning
        & 19 & 10.3\% \\

        Data Analysis
        & 15 & 8.2\% \\

        Multimodal Processing
        & 12 & 6.5\% \\

        Programming \& Debugging
        & 3 & 1.6\% \\

        Text Processing
        & 2 & 1.1\% \\
        \midrule

        \textbf{Total}
        & \textbf{184} & \textbf{100\%} \\

        \bottomrule
    \end{tabular*}
\end{table}

\subsection{Root Cause Taxonomy}

Table~\ref{tab:failure_taxonomy} classifies 184 failure
instances into nine categories.
Errors are distributed across \textit{structural}
(e.g., Code Error) and \textit{semantic}
(e.g., Hallucination) domains, empirically justifying
the \textsc{DuoTrace} dual-view architecture.

\begin{table}[t]
    \centering
    \caption{Root Cause Classification of 184 Failure Instances.}
    \label{tab:failure_taxonomy}

    \fontsize{8pt}{9.6pt}\selectfont
    \setlength{\tabcolsep}{2pt}
    \renewcommand{\arraystretch}{1.15}

    \begin{tabular*}{\columnwidth}{
        @{\extracolsep{\fill}} l c c @{}
    }
        \toprule
        \textbf{Failure Category}
        & \textbf{Count}
        & \textbf{Ratio} \\
        \midrule

        Code Error
        & 40 & 21.7\% \\

        Insufficient Information
        & 25 & 13.6\% \\

        Logic Error
        & 22 & 12.0\% \\

        Factual Error
        & 21 & 11.4\% \\

        Understanding Error
        & 20 & 10.9\% \\

        Tool Misuse
        & 19 & 10.3\% \\

        Model Hallucination
        & 17 & 9.2\% \\

        Validation Failure
        & 14 & 7.6\% \\

        Resource Limitation
        & 6 & 3.3\% \\
        \midrule

        \textbf{Total}
        & \textbf{184} & \textbf{100\%} \\

        \bottomrule
    \end{tabular*}
\end{table}

% ---------------------------------------------------------
% Section A: Implementation Details
% ---------------------------------------------------------
\section{Implementation Details}
\label{sec:appendix_implementation}

To facilitate full reproducibility of our experiments, we provide the comprehensive implementation details of the \textsc{DuoTrace} framework below.

\textbf{Model Architecture.}
We use Sentence Transformer (\texttt{all-MiniLM-L6-v2})
to obtain $384$-dimensional embeddings of agent identities,
operations, and node content.
A pretrained text-pair encoder $\rho$ encodes query--content
pairs for task alignment and ancestor--content pairs for
historical relations. We average the ancestor-pair
representations for each node and use a zero vector for
the root.
We use the formulation in Eq.~\eqref{eq:main_definition}.
Here, $P_s$, $P_q$, and $P_r$ are learned affine
projections, and $\oplus$ denotes concatenation.
The fused contextual representation is combined with
the agent and operation embeddings and projected to
a $128$-dimensional node input.
A Child-Sum Tree-LSTM with $128$-dimensional hidden states
aggregates information from child nodes along invocation
edges. Each node's hidden state parameterizes a
$64$-dimensional diagonal Gaussian posterior.

The decoder uses MLPs with ReLU activations to reconstruct
the five feature blocks: agent identity, operation,
content, task relation, and history relation.
Each reconstruction output matches the dimensionality
of its corresponding target.
A separate prediction head outputs the Gaussian mean
and log-variance of node-level token consumption.
The root's history-relation is excluded from the
reconstruction loss.

\textbf{Training Hyperparameters.} 
The \textsc{DuoTrace} module is trained end-to-end using the standard Adam optimizer. The learning rate is initialized to $2 \times 10^{-3}$. We train the model for $50$ epochs across all datasets. During the joint optimization, the KL-divergence weight ($\beta$) in the ELBO objective is set to a constant value of $0.01$. The reconstruction loss for the semantic embeddings is calculated using Mean Squared Error (MSE), while the token consumption is optimized via Negative Log-Likelihood (NLL).

\textbf{Inference and Evaluation Setup.}
At inference time, we decode each node's posterior mean
and compute its anomaly score from the weighted feature
reconstruction errors and token-consumption NLL.
We standardize these scores within each trajectory
and rank nodes by
$R(v_i)=Z(v_i)+\alpha\max(0,Z(v_i)-Z(\pi(v_i)))$,
where $\pi(v_i)$ denotes the parent node; the root's
mutation term is set to zero.
We select the top 20\% of nodes and retain each candidate's
parent and direct children. Overlapping context is
deduplicated before the retained evidence is passed
to the downstream attribution method.

\textbf{Computational Infrastructure.} 
All models and experiments are implemented in PyTorch on a single NVIDIA RTX 3060 GPU with 12GB of VRAM. Training the \textsc{DuoTrace} module from scratch requires only a few hours, underscoring its computational efficiency.

% --------------------------------------------------------
% Section D: Counterfactual Alignment Analysis
% ---------------------------------------------------------

\section{Generalization to Held-Out Task Categories}
\label{app:heldout_task}

We train the anomaly detector on Information Retrieval,
Programming \& Debugging, and Text Processing, and evaluate
it on Math \& Logic Reasoning, Data Analysis, and Multimodal
Processing without target-category retraining.
\textsc{DuoTrace} improves both agent and step accuracy
across all four downstream methods
(Table~\ref{tab:heldout_task}), supporting transfer
to task categories unseen during detector training.

\begin{table}[t]
  \centering
  \caption{Attribution performance on held-out task categories
  without target-category retraining.
  Bold indicates \textsc{DuoTrace}-enhanced results.}
  \label{tab:heldout_task}

  \fontsize{8pt}{9.6pt}\selectfont
  \setlength{\tabcolsep}{2pt}
  \renewcommand{\arraystretch}{1.15}

  \begin{tabularx}{\columnwidth}{
    @{} l >{\raggedright\arraybackslash}X c c @{}
  }
    \toprule
    \textbf{Type}
    & \textbf{Method}
    & \textbf{Agent}
    & \textbf{Step} \\
    \midrule

    \multirow{4}{*}{\shortstack[l]{LLM-based\\Prompting}}
    & GPT-5
    & 34.8 & 8.7 \\

    & \textbf{GPT-5 + \textsc{DuoTrace}}
    & \textbf{41.3} & \textbf{15.2} \\

    & Qwen3.5-plus
    & 30.4 & 6.5 \\

    & \textbf{Qwen3.5-plus + \textsc{DuoTrace}}
    & \textbf{37.0} & \textbf{13.0} \\
    \midrule

    \multirow{4}{*}{\shortstack[l]{LLM-as-\\a-Judge}}
    & ECHO
    & 41.3 & 17.4 \\

    & \textbf{ECHO + \textsc{DuoTrace}}
    & \textbf{52.2} & \textbf{23.9} \\

    & CORRECT
    & 45.7 & 21.7 \\

    & \textbf{CORRECT + \textsc{DuoTrace}}
    & \textbf{54.3} & \textbf{28.3} \\

    \bottomrule
  \end{tabularx}
\end{table}

\end{document}